\pdfoutput=1
\documentclass[11pt]{article}
\usepackage[preprint]{acl}
\usepackage{times,latexsym}
\usepackage[T1]{fontenc}
\usepackage[utf8]{inputenc}
\usepackage{microtype,inconsolata}
\usepackage{graphicx,booktabs,amsmath,amssymb}
\usepackage{multirow}
\usepackage{tikz}
\usetikzlibrary{arrows.meta,positioning}
\title{A Token Is Not a Unit of Difficulty: Segmentation Controls Where Adaptive Computation Goes Inside a Word}
\author{Dipto Sumit, Mahima Mahzabin Sandria, Farig Sadeque, Md. Golam Rabiul Alam \\
  BRAC University, Dhaka, Bangladesh \\
  \texttt{dipto.sumit@g.bracu.ac.bd}}

\begin{document}
\maketitle
\begin{abstract}
Early-exit language models decide for each token how much computation to spend, usually by stopping once the prediction is confident. Yet a token is not a fixed unit: tokenizers cut words into different numbers of pieces in different languages. We ask where inside a word such a policy spends its budget. In Qwen3, along two adaptive axes (the timesteps of a spiking conversion and the layers of the original Transformer), confidence stopping gives the continuation pieces of a word less computation than its first piece in English, German, and Turkish, but more in Bangla. Segmentation drives this. Re-segmenting the same parallel passages with BPE-dropout shifts the gap by up to eight exits on both axes, small models trained under two tokenizers shift Bangla's gap by six exits, and BLOOM and mGPT, which split Bangla less finely, give it the English pattern. Characters per token do not predict the gap across languages; within-word surprisal explains 75\% of its variance. A learned stopping head beats confidence stopping by 7 to 8 points, but gains less than one point over the best mix of fixed exits at equal cost, and sometimes loses. All results come from offline analysis, not measured speedups.
\end{abstract}

\section{Introduction}
\begin{figure*}[t]
\centering
\begin{tikzpicture}[font=\scriptsize, x=1cm, y=1cm,
  box/.style={draw=black!40, rounded corners=2pt, fill=white, align=left,
              inner sep=3pt, text width=#1, anchor=north west},
  box/.default=3.22cm,
  stage/.style={fill=black!6, rounded corners=4pt},
  num/.style={circle, fill=black!75, text=white, font=\scriptsize\bfseries, inner sep=1.2pt},
  hdr/.style={font=\footnotesize\bfseries, anchor=west},
  arr/.style={-{Stealth[length=5pt,width=5pt]}, line width=1.4pt, black!45}]
\definecolor{spk}{HTML}{0072B2}
\definecolor{lyr}{HTML}{009E73}
\def\top{-0.4}
\def\bot{-4.12}
\foreach \x in {0,4.12,8.24,12.36}{\fill[stage] (\x-0.08,0.3) rectangle (\x+3.55,\bot);}
\foreach \x in {3.55,7.67,11.79}{\draw[arr] (\x+0.03,-1.9) -- (\x+0.46,-1.9);}
\foreach \x/\n/\t in {0/1/Segment the text, 4.12/2/Compute every exit,
                      8.24/3/Choose exits, 12.36/4/Measure}{
  \node[num] at (\x+0.12,0.02) {\n};
  \node[hdr] at (\x+0.3,0.02) {\t};
}
\node[box] (s1a) at (0,\top) {\textbf{Canonical tokenizers}\\
  Qwen3, BLOOM, and mGPT (\S\ref{sec:families})};
\node[box] (s1b) at ([yshift=-0.14cm]s1a.south west) {\textbf{BPE-dropout ladder}\\
  rates 0 to .85; same passages, model, and vocabulary (\S\ref{sec:reversals})};
\node[box] (s1c) at ([yshift=-0.14cm]s1b.south west) {\textbf{Two tokenizers, from scratch}\\
  differ only in Bangla coverage (\S\ref{sec:controlled})};
\node[box=2.45cm] (s2a) at (4.12,\top) {\textbf{Timestep axis}\\
  Qwen3 converted to a partly spiking network; exit $=$ timestep};
\node[box=2.45cm] (s2b) at ([yshift=-0.3cm]s2a.south west) {\textbf{Depth axis}\\
  original model read by frozen tuned lenses; exit $=$ layer};
\begin{scope}[shift={(6.95,-1.15)}]
  \draw[black!40] (0,0) -- (0.5,0);
  \foreach \s in {0.04,0.15,0.2,0.33,0.45}{\draw[spk, line width=0.9pt] (\s,0) -- (\s,0.4);}
  \node[font=\tiny, black!60, anchor=north] at (0.25,-0.03) {16 steps};
\end{scope}
\begin{scope}[shift={(6.92,-2.95)}]
  \foreach \k in {0,...,4}{
    \fill[lyr!55] (0,{\k*0.2}) rectangle (0.32,{\k*0.2+0.13});
    \draw[-{Stealth[length=2pt]}, black!50] (0.35,{\k*0.2+0.065}) -- (0.5,{\k*0.2+0.065});
    \fill[black!60] (0.54,{\k*0.2+0.065}) circle (0.03);
  }
  \node[font=\tiny, black!60, anchor=north] at (0.28,-0.03) {16 exits};
\end{scope}
\node[box] (s3a) at (8.24,\top) {\textbf{Cached trajectories}\\
  confidence $c_{i,t}$ and loss $\ell_{i,t}$ of every token at every exit, teacher-forced};
\node[box] (s3b) at ([yshift=-0.14cm]s3a.south west) {\textbf{Policies at an equal budget}\\
  confidence threshold $\tau$\\
  learned stopping head (\S\ref{sec:head})\\
  best mix of two fixed exits};
\node[box] (s4a) at (12.36,\top) {\textbf{Allocation gap} $G=\bar t_C-\bar t_I$\\
  continuations vs.\ word starts, and what predicts it (\S\ref{sec:failed})};
\node[box] (s4b) at ([yshift=-0.14cm]s4a.south west) {\textbf{Demand and service}\\
  \mbox{$G=G_{\text{dem}}+G_{\text{svc}}$}, what tokens need vs.\ what the policy adds (\S\ref{sec:service})};
\node[box] (s4c) at ([yshift=-0.14cm]s4b.south west) {\textbf{Accuracy at equal cost}\\
  head vs.\ confidence vs.\ fixed exits (\S\ref{sec:head})};
\end{tikzpicture}
\caption{Experimental design, and where each part is reported. (1) We change segmentation in three ways: across the tokenizers of pretrained models, with BPE-dropout on a fixed model, and by training small models under two tokenizers. (2) Each model gives a prediction at every exit along one of two adaptive axes. (3) Stopping policies choose one exit per token at the same average budget. (4) We compare the exits that word starts and continuations receive, split that gap into what tokens need and what the policy adds, and compare policies at equal cost. Icons are schematic.}
\label{fig:pipeline}
\end{figure*}
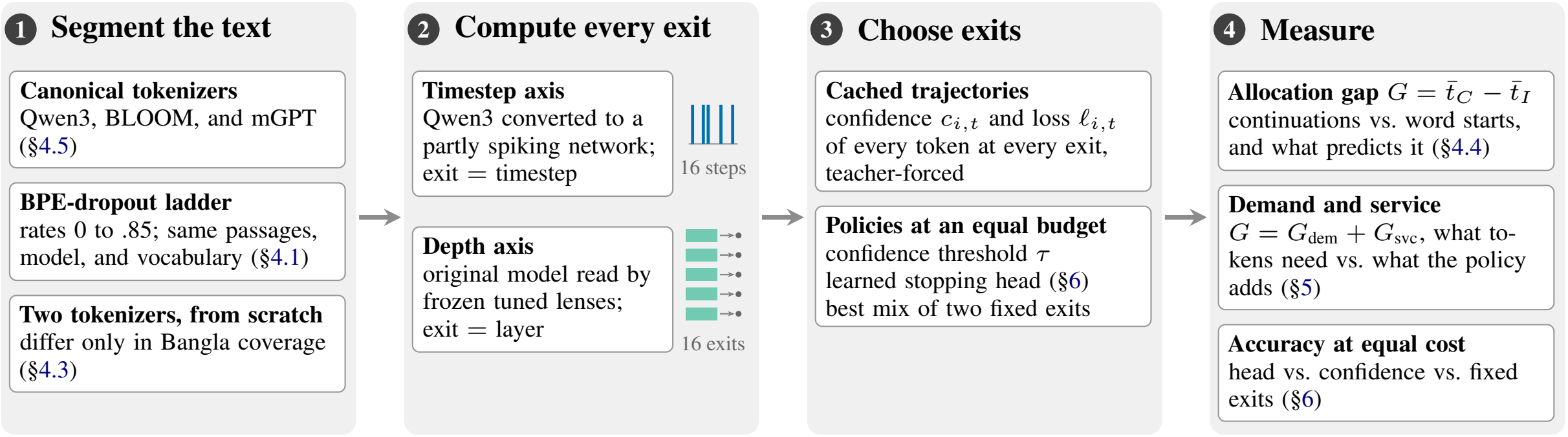

Early-exit and other adaptive-computation methods let a language model spend more work on hard predictions and stop early on easy ones \citep{depth,calm}. The decision is made one token at a time, often by stopping as soon as the model's confidence passes a threshold, so the token becomes the unit of computation. Tokens, however, are not comparable units across languages. A subword tokenizer may keep a common English word whole, split a Turkish word into several reusable pieces, and break a Bangla word into fragments smaller than one character. A stopping rule that works per token inherits these choices.

\begin{figure}[t]
\centering
\begin{tikzpicture}[font=\footnotesize, x=1cm, y=1cm]
\definecolor{wstart}{HTML}{0072B2}
\definecolor{wcont}{HTML}{E69F00}
\node[anchor=west] at (-0.45,2.0) {\textbf{(a)} Each piece stops at its own exit};
\def\cw{0.92}\def\ch{0.25}
\foreach \tok/\cls/\ex [count=\k from 0] in
  {{\textvisiblespace The}/1/2, {\textvisiblespace token}/1/6, {izer}/0/2,
   {\textvisiblespace split}/1/5, {s}/0/1, {\textvisiblespace word}/1/4, {s}/0/1}{
  \pgfmathsetmacro{\x}{\k*\cw}
  \foreach \r in {1,...,6}{
    \draw[black!20, fill=white] (\x+0.05,{(\r-1)*\ch}) rectangle ++(\cw-0.1,\ch-0.04);
  }
  \ifnum\cls=1 \def\col{wstart}\else\def\col{wcont}\fi
  \foreach \r in {1,...,\ex}{
    \fill[\col!75] (\x+0.05,{(\r-1)*\ch}) rectangle ++(\cw-0.1,\ch-0.04);
  }
  \node[anchor=north, font=\ttfamily\footnotesize] at (\x+\cw/2,-0.04) {\tok};
}
\draw[-{Stealth[length=4pt]}] (-0.1,0) -- (-0.1,1.6);
\node[rotate=90, anchor=south] at (-0.12,0.75) {exits used};
\fill[wstart!75] (0.05,-0.74) rectangle ++(0.25,0.17);
\node[anchor=west] at (0.32,-0.655) {word start $I$};
\fill[wcont!75] (2.3,-0.74) rectangle ++(0.25,0.17);
\node[anchor=west] at (2.57,-0.655) {continuation $C$};
\node[anchor=west] at (0.0,-1.08) {$G=\bar t_C-\bar t_I=1.33-4.25=-2.92$};
\begin{scope}[yshift=-5.2cm]
\node[anchor=west] at (-0.45,3.55) {\textbf{(b)} What a token gets vs.\ what it needs};
\def\dx{0.76}
\begin{scope}[yshift=1.7cm]
  \draw[black!40] (0,0) -- (8*\dx,0);
  \draw[black!40] (0,0) -- (0,1.1);
  \draw[dashed, black!60] (0,0.78) -- (8*\dx,0.78) node[right] {$\tau$};
  \draw[black!80, thick, mark=*, mark size=1.1pt] plot coordinates
    {(1*\dx,0.10) (2*\dx,0.19) (3*\dx,0.31) (4*\dx,0.47) (5*\dx,0.66)
     (6*\dx,0.82) (7*\dx,0.89) (8*\dx,0.93)};
  \node[anchor=west] at (0.05,1.02) {confidence $c_{i,t}$};
\end{scope}
\fill[wcont!25] (0,0.14) rectangle (8*\dx,0.31);
\node[anchor=west, font=\scriptsize] at (0.06,0.225) {within $\varepsilon$ of final loss};
\draw[black!40] (0,0) -- (8*\dx,0);
\draw[black!40] (0,0) -- (0,1.4);
\draw[black!80, thick, mark=*, mark size=1.1pt] plot coordinates
  {(1*\dx,1.35) (2*\dx,0.92) (3*\dx,0.55) (4*\dx,0.29) (5*\dx,0.26)
   (6*\dx,0.23) (7*\dx,0.21) (8*\dx,0.19)};
\node[anchor=west] at (0.9,1.3) {loss $\ell_{i,t}$};
\draw[densely dotted, black!70] (4*\dx,0) -- (4*\dx,2.95);
\draw[densely dotted, black!70] (6*\dx,0) -- (6*\dx,2.95);
\node[anchor=south, inner sep=1pt] at (4*\dx,2.95) {demand $d_i$};
\node[anchor=south, inner sep=1pt] at (6*\dx,2.95) {exit $t_i$};
\foreach \t in {1,...,8}{\node[anchor=north, font=\scriptsize, black!60] at (\t*\dx,0) {\t};}
\node[anchor=north west, font=\scriptsize, black!60] at (8*\dx+0.04,0) {$t$};
\draw[{Stealth[length=4pt]}-{Stealth[length=4pt]}] (4*\dx,-0.45) -- (6*\dx,-0.45)
  node[midway, below] {service $s_i=t_i-d_i=+2$};
\end{scope}
\end{tikzpicture}
\caption{How we measure allocation. This is a schematic: the segmentation and values are illustrative, not model output. (a) Confidence stopping gives each token its own exit. Pieces that begin with the tokenizer's space marker are word starts ($I$), the rest are continuations ($C$), and the allocation gap $G$ is the difference between their mean exits. An exit is a spiking timestep on one axis and a layer read through a tuned lens on the other. (b) For one token, the exit $t_i$ is where confidence first reaches the threshold $\tau$, and the demand $d_i$ is where the loss first comes within $\varepsilon$ of its final value. Service $s_i=t_i-d_i$ is positive when the policy keeps computing after the token is done.}
\label{fig:overview}
\end{figure}
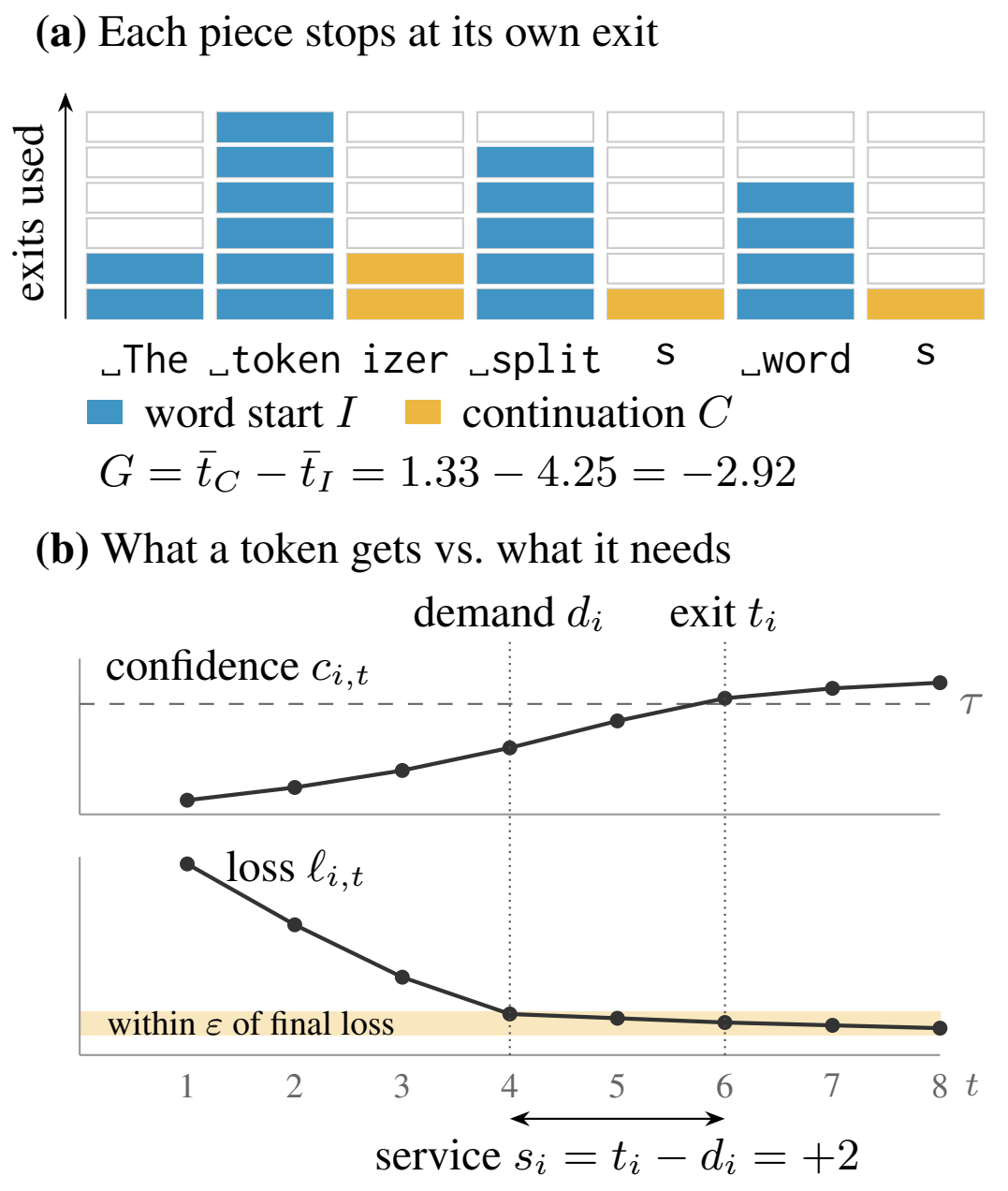

Earlier work showed that tokenizers make some languages longer and more expensive to process \citep{fairness,cost}. We ask a different question. If a stopping rule has the same average budget in every language, where inside a word does that budget go? We divide tokens into \emph{word starts}, which begin with the tokenizer's space marker, and \emph{continuations}, and measure the \emph{allocation gap}: how many more exits continuation pieces receive than word starts (Figure~\ref{fig:overview}a).

The sign of this gap depends on the language. In English, German, and Turkish, confidence stopping gives continuation pieces less computation than word starts; in Bangla it gives them more. The pattern appears on two different adaptive axes of the same Qwen3 models: the timesteps of a version converted into a spiking network, and the layers of the original Transformer read through tuned lenses \citep{lens}.

Three experiments, summarized in Figure~\ref{fig:pipeline}, point to segmentation as the cause. First, re-segmenting the same parallel passages with BPE-dropout \citep{bpedropout}, with model, vocabulary, and readout fixed, turns the English, German, and Turkish gaps from negative to positive on both axes. Second, small models trained from scratch under two tokenizers that differ only in their Bangla coverage show a six-exit difference in Bangla's gap, while English does not move. Third, BLOOM and mGPT, whose tokenizers split Bangla less finely, give Bangla the same negative gap as English. Bangla's positive gap in Qwen3 belongs to how Qwen3 segments it.

Characters per token still do not explain why languages differ. Canonical German is segmented about as finely as English under mild dropout, yet the two gaps have opposite signs. A better predictor is how much harder continuation pieces are to predict than word starts at full computation; this surprisal contrast explains 75\% of the variance across 15 languages.

A gap in allocation does not show that the policy is wrong, because word starts may simply need more computation. We therefore measure each token's \emph{demand}, the first exit at which its loss is already close to final (Figure~\ref{fig:overview}b). For the pretrained models on the timestep axis, much of the gap, including the Bangla reversal, is demand. On the depth axis and in the small from-scratch models it is not. Finally, a small learned stopping head beats confidence stopping by 7 to 8 points of token accuracy, but gains less than a point over the best mixture of fixed exits at the same cost.

All experiments use cached, teacher-forced trajectories. Our results describe stopping policies and the predictions they select; we do not measure real speedups.\footnote{Code is available at \url{https://anonymous.4open.science/r/spikelm_naacl-820C/}.}

\section{Related Work}
\paragraph{Tokenization across languages.}
Subword vocabularies trade vocabulary size against sequence length \citep{bpe}. This trade-off falls differently on different languages, which affects cost, context length, and access for their users \citep{cost,fairness}. Tokenization also distorts distances between positions \citep{distance}, which matters here because re-segmenting text moves every token's position. We use an existing method, BPE-dropout, as a probe and do not propose a new tokenizer.

\paragraph{Adaptive depth and early exit.}
The Depth-Adaptive Transformer learns how many layers to use for each token \citep{depth}. CALM calibrates confidence-based exits and deals with the hidden states that early exits leave missing \citep{calm}, and LayerSkip trains intermediate exits for self-speculative decoding \citep{layerskip}. Other systems make early exit practical in deployment, with a fixed exit schedule per position that works with batching and key-value caches \citep{skipdecode}, learned routers at checkpoint layers \citep{tide}, or generated cache entries for skipped layers \citep{adept}. These systems measure real speedups, which we do not. Our question comes before deployment: which tokens a stopping rule favours, and whether the tokenizer decides this. A learned router could carry the same bias, and our measurements apply to it unchanged. \citet{eefp} show that better-calibrated exits do not by themselves give better adaptive computation, which fits our finding that confidence is a weak allocator. Adaptive computation has also been used to learn character grouping in translation \citep{learnsegment}, so linking segmentation and computation is not new. We instead change the segmentation seen by a fixed model, or the tokenizer a model is trained with, and look at the effect within words.

\paragraph{Spiking models and intermediate readouts.}
Converting a trained network to a spiking one requires controlling the error of simulating it for a finite number of timesteps \citep{qcfs}, and dynamic-timestep spiking networks choose that number per input \citep{dtsnn}. Our converted model is a hybrid in which only one activation per block spikes. For the depth axis we use affine tuned lenses, which read intermediate layers far better than an untrained logit lens \citep{lens}. A tuned lens is itself trained, so we test how much our results depend on it.

\section{Setup}
\subsection{What we measure}

For target token $y_i$ and exit $t\in\{1,\ldots,T\}$, let $p_{i,t}$ be the predicted distribution and $c_{i,t}=\max_v p_{i,t}(v)$ its confidence. Confidence stopping exits at the first step whose confidence reaches a threshold $\tau$:
\begin{equation}
 t_i(\tau)=\min\bigl(\{t:c_{i,t}\geq\tau\}\cup\{T\}\bigr).
\end{equation}
We set $\tau$ so that the mean exit $\bar t$ meets a target budget, and report next-token accuracy at the chosen exits. An exit is a spiking timestep on one axis and one of 16 sampled layers on the other, and neither is a FLOP count.

A token is a \emph{word start} ($I$) if its vocabulary entry begins with the byte-BPE space or newline marker, and a \emph{continuation} ($C$) otherwise. These are tokenizer classes, not linguistic word boundaries. The allocation gap is
\begin{equation}
 G=\mathbb{E}[t_i\mid i\in C]-\mathbb{E}[t_i\mid i\in I],
 \label{eq:gap}
\end{equation}
so a positive gap means continuation pieces get more computation.

Different groups may need different amounts of computation, so we also measure need. With $\ell_{i,t}$ the loss on the target, a token's \emph{demand} is the first exit at which the loss comes within $\varepsilon=0.05$ nats of its final value,
\begin{equation}
 d_i=\min\bigl(\{t:\ell_{i,t}\leq\ell_{i,T}+\varepsilon\}\cup\{T\}\bigr),
 \label{eq:demand}
\end{equation}
and its \emph{service} is $s_i=t_i-d_i$, positive when the policy keeps computing after the token is done and negative when it stops too soon. Demand and service gaps are formed as in Equation~\ref{eq:gap} and add up exactly, $G=G_{\text{dem}}+G_{\text{svc}}$. Demand comes from the same model whose policy it judges. It asks whether the policy spends where this model still improves, not how hard a word is in any absolute sense. The same demand is the training label of the stopping head in Section~\ref{sec:head}. Appendix~\ref{app:criterion} repeats the analysis with a stricter target, under which the loss must stay within $\varepsilon$, and Appendix~\ref{app:eps} varies $\varepsilon$.

Finally, the surprisal contrast
\begin{equation}
 P=\mathbb{E}[-\log p_{i,T}(y_i)\mid C]
 -\mathbb{E}[-\log p_{i,T}(y_i)\mid I]
 \label{eq:surprisal}
\end{equation}
measures how much harder continuation pieces are to predict than word starts at full computation.

\subsection{Models and the two adaptive axes}
We use Qwen3-0.6B and Qwen3-1.7B \citep{qwen} with their original vocabulary and weights, and give each model two adaptive axes. On the \emph{timestep axis}, the model is converted into a partly spiking network. In each of its 28 blocks, integrate-and-fire units encode the input to the MLP down-projection while everything else stays analog. The model runs for $T=16$ timesteps, and each timestep's prediction uses the model's own output layer on logits averaged over the steps so far. Neuron thresholds are calibrated once per channel, at the 70th percentile of absolute activations on eight WikiText blocks (Appendix~\ref{app:conversion}). No readout is fitted, which makes this our main instrument.

On the \emph{depth axis}, we use the unconverted model with affine tuned lenses \citep{lens} at 15 intermediate layers, plus the real output as the 16th exit (Appendix~\ref{app:lens}). This is the axis most early-exit systems use. We make no claim about energy savings on neuromorphic or other hardware. To test whether the results are specific to Qwen3, we also run the depth probe on BLOOM-1.1B \citep{bloom} and mGPT-1.3B \citep{mgpt}, which split Bangla into about 4 and 1.4 characters per token, compared with 0.94 for Qwen3.

\subsection{Data and segmentation interventions}
We evaluate next-token prediction on WikiText-103 \citep{wikitext}, English and Bangla passages from Belebele \citep{belebele}, and Bangla text from Sangraha \citep{indic}. Belebele serves only as a source of parallel text, so all accuracies are token accuracies. Standard runs use 48 blocks of 512 tokens (32 for Sangraha). A wider sweep covers 18 Belebele languages, and our cross-language regression uses the 15 that separate words with spaces. In these runs each language's passages are sampled independently, so they are not translations of each other.

To change segmentation without changing the model, we apply BPE-dropout \citep{bpedropout} to English, German, Turkish, and Bangla at rates 0, .1, .2, .3, .5, .7, and .85, with three random segmentations at each nonzero rate. The model, vocabulary, and conversion thresholds stay fixed, and the realized token sequences are archived. Each cell gets its own threshold for the target budget, fitted without labels. Intervals resample passages or blocks, and the spread across the three draws reflects segmentation randomness. Noncanonical segmentations are still unusual inputs for a pretrained model, which is why Section~\ref{sec:controlled} also trains models under different tokenizers.

\begin{figure*}[t]
\centering
\includegraphics[width=\textwidth]{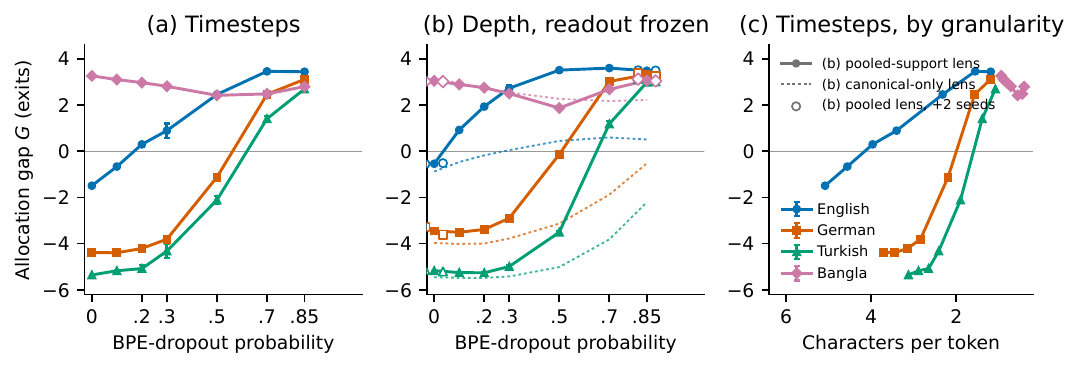}
\caption{Re-segmenting the same 60 parallel passages with BPE-dropout, at a budget of four exits. Every panel uses the same realized tokens. Points are means over three segmentation draws, and error bars show one standard deviation. (a) Spiking timesteps: English, German, and Turkish change sign, while Bangla starts positive and stays positive. (b) Layers of the unconverted model, read through one tuned lens per language that is fitted once and then frozen. Solid lines use a lens fitted on both canonical and fragmented text, dotted lines a lens fitted on canonical text only, and hollow markers two further lens seeds. (c) The timestep curves against characters per token. At equal granularity the languages sit several exits apart.}
\label{fig:ladder}
\end{figure*}

\section{Segmentation Changes Allocation}
\subsection{Fragmenting words reverses the gap}
\label{sec:reversals}
\label{sec:aligned}
Our main design uses 60 Belebele passages that exist in all four languages. Each passage is scored as its own sequence, with an end-of-sequence marker in front so that every character is a target. The scored text is then identical at every dropout rate, and the four languages are translations of the same passages.

Figure~\ref{fig:ladder}a shows the timestep axis. As words are broken into smaller pieces, English moves from $G=-1.49$ to $+3.44$, German from $-4.38$ to $+3.11$, and Turkish from $-5.35$ to $+2.69$. Each crosses zero. Bangla, already split below the character level, starts at $+3.26$ and stays positive. These changes are far larger than the variation between draws. The median standard deviation across draws is 0.06 timesteps, and the only rung whose draws disagree in sign is German at .5, where the mean is $-0.04$. A second ladder that cuts concatenated text into fixed 512-token blocks gives the same signs and similar sizes (Appendix~\ref{app:ladder}), and removing tokens that consist only of a space changes the canonical gaps by at most 0.03.

\subsection{The same response on the depth axis}
\label{sec:depthladder}
If the effect belongs to confidence stopping rather than to the spiking conversion, it should also appear on depth. We scored the same token sequences, in all 76 cells, through the unconverted 0.6B model. Each language has one tuned lens that is fitted once and then frozen for every rung, so the readout cannot adapt to the treatment. This lens is fitted on held-out passages segmented at dropout 0, .3, and .85, so it has seen fragmented input.

The depth curves closely follow the timestep curves (Figure~\ref{fig:ladder}b). English moves from $-0.54$ to $+3.48$, German from $-3.45$ to $+3.32$, and Turkish from $-5.17$ to $+2.98$, while Bangla stays positive. The three ranges, 4.1, 6.8, and 8.2 exits, are close to the 5.0, 7.5, and 8.1 exits on timesteps, and two further lens seeds move no endpoint by more than 0.17. A lens fitted on canonical text alone shows the same direction at 30--45\% of the size because it reads fragmented input poorly: at dropout .85, English word starts have a first-exit loss of 9.1 nats through that lens and 2.2 through the pooled one. The way a confidence policy responds to segmentation is therefore not a property of the spiking axis.

\subsection{Models trained under two tokenizers}
\label{sec:controlled}
So far a fixed model reads text segmented in ways it never saw in training, so part of the effect could come from that mismatch. To remove it, we trained small models from scratch under two tokenizers that differ only in how much Bangla text they were built from. Both are byte-level BPE with 16{,}000 pieces. Arm A saw 4{,}000 Bangla characters and splits Bangla at $0.93$ characters per token, close to Qwen3's $0.94$. Arm B saw 10 million and splits it at $1.40$. English barely changes ($1.01\times$). For each arm we trained a 49.1M-parameter decoder on the same documents in the same order, with three paired seeds, and read it through per-language lenses that are fitted once and frozen. We evaluate at a budget of 6.5 of 13 exits on 300 parallel Belebele passages (Appendix~\ref{app:controlled}).

The coarser Bangla tokenizer moves Bangla's gap by $-6.11$ exits on average, from $+0.86$ to $-5.25$, and every seed lies between $-6.00$ and $-6.27$. English moves by only $-0.02$, with seeds of both signs. Comparing checkpoints after an equal number of training tokens gives the same $-6.11$. Arm A reproduces the pretrained pattern (English negative, Bangla positive) in models that never see a noncanonical sequence, and the direction matches the ladders: coarser segmentation gives a more negative gap.

One result does not carry over. In these models the tokenizer changes service, not demand. The stable demand gap moves by only $-0.39$ to $-0.57$ exits, under a tenth of the change in allocation in every seed, and a 125.6M-parameter pair shows the same split at two budgets. The small models hardly differentiate demand within words at all ($|G_{\text{dem}}|\le1.1$, against $+3.9$ to $+4.2$ for the pretrained model). The main remaining difference is training: 386M tokens, against the trillions seen by Qwen3.

\subsection{Granularity does not predict the gap across languages}
\label{sec:failed}
\label{sec:info}
If the gap depended only on how finely text is segmented, languages with similar characters per token would have similar gaps. Before running the German and Turkish ladders, we used the English ladder to predict their canonical gaps from granularity: about $+0.7$ to $+0.9$ for German (3.60 characters per token) and $+1.5$ to $+1.8$ for Turkish (3.06). The block-based ladder then gave $-4.00$ and $-5.03$. Canonical German and English at dropout .2 are segmented almost equally finely (3.60 and 3.72 characters per token) but have gaps of opposite sign, and Figure~\ref{fig:ladder}c shows the four curves offset by several exits at equal granularity.

The surprisal contrast $P$ predicts the gap much better. Across 15 languages it explains $R^2=.746$ of the variance in $G$, against .351 for log characters per token. Given $P$, granularity adds a partial $R^2$ of only .070; given granularity, $P$ adds .636 (Appendix~\ref{app:convergence}). A shared slope in granularity with a separate intercept per language fits much better than a single line (in-sample $R^2$ .877 against .386), which again says that each language has its own baseline. Both $G$ and $P$ come from the same model, so this describes the model's behaviour and does not identify a linguistic cause.

\subsection{Other model families}
\label{sec:families}
\begin{figure}[t]
\centering
\includegraphics[width=\columnwidth]{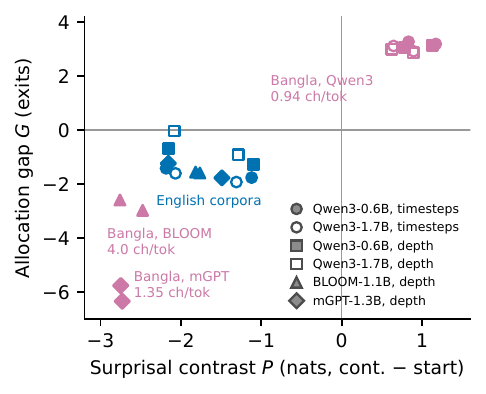}
\caption{Allocation gap against surprisal contrast for all 24 canonical corpus--model pairs: two Qwen3 sizes on both axes, and BLOOM-1.1B and mGPT-1.3B on depth. Filled markers are 0.6B or the other families, hollow markers 1.7B; blue marks English corpora and pink Bangla ones. Every pair whose gap differs from zero lies in the upper-right or lower-left quadrant. Where Bangla falls depends on how finely its tokenizer splits it, and not monotonically.}
\label{fig:families}
\end{figure}

If Bangla's positive gap comes from Qwen3's fine segmentation, a model whose tokenizer splits Bangla more coarsely should give Bangla an English-like gap. We ran the same per-corpus tuned-lens probe on BLOOM-1.1B, which splits Bangla at about 4 characters per token (40\% continuation pieces, against 85\% in Qwen3), and on mGPT-1.3B, at about 1.35. Both give Bangla a negative gap (Figure~\ref{fig:families}): $-2.59$ and $-2.97$ under BLOOM, and $-6.34$ and $-5.76$ under mGPT. English is negative in both models too ($-1.56$ and $-1.59$ under BLOOM, $-1.24$ and $-1.77$ under mGPT), as is every surprisal contrast. mGPT's Bangla gap is within about one exit of Arm B's ($-5.25$) at nearly the same granularity, although one model is pretrained and the other was trained by us. The gap is not monotone in granularity: 0.94, 1.35, and 4.0 characters per token give about $+3$, $-6$, and $-3$.

This is also an out-of-family test of the surprisal contrast. The signs of $G$ and $P$ agree in 23 of the 24 canonical corpus--model pairs, and the exception, WikiText on the 1.7B depth probe, has a gap of $-0.04$ that is indistinguishable from zero. That case also shows that the English side of the pattern is weaker on depth than on timesteps. All of these are canonical measurements; we did not run the ladders on BLOOM or mGPT.

\section{Is the Gap What the Model Needs?}
\label{sec:service}
\begin{table}[t]
\centering\small\setlength{\tabcolsep}{4pt}
\begin{tabular}{llrrr}
\toprule
Setting & Corpus & $G_{\text{dem}}$ & $G$ & $G_{\text{svc}}$\\
\midrule
\multirow{4}{*}{Time / 0.6B}
 & WikiText & $-1.89$ & $-1.42$ & $+0.47^{*}$\\
 & Belebele-en & $-1.43$ & $-1.76$ & $-0.33^{*}$\\
 & Belebele-bn & $+2.51$ & $+3.19$ & $+0.68^{*}$\\
 & Sangraha-bn & $+2.52$ & $+3.27$ & $+0.75^{*}$\\
\midrule
\multirow{4}{*}{Time / 1.7B}
 & WikiText & $-1.56$ & $-1.60$ & $-0.05$\\
 & Belebele-en & $-1.86$ & $-1.93$ & $-0.07$\\
 & Belebele-bn & $+1.40$ & $+2.92$ & $+1.52^{*}$\\
 & Sangraha-bn & $+1.53$ & $+3.12$ & $+1.59^{*}$\\
\midrule
\multirow{4}{*}{Depth / 0.6B}
 & WikiText & $+0.68$ & $-0.70$ & $-1.38^{*}$\\
 & Belebele-en & $-0.60$ & $-1.26$ & $-0.66^{*}$\\
 & Belebele-bn & $+3.28$ & $+3.15$ & $-0.14$\\
 & Sangraha-bn & $+3.38$ & $+3.06$ & $-0.32^{*}$\\
\bottomrule
\end{tabular}
\caption{Allocation gap $G$ split into demand $G_{\text{dem}}$ and service $G_{\text{svc}}$, in exits, at a budget of four and canonical segmentation. Asterisks mark service gaps whose cluster-bootstrap interval excludes zero. Demand has the sign of $G$ in eleven of twelve settings and makes up roughly half of it or more in each of them. The exception, WikiText on depth, disappears under the stable target of Appendix~\ref{app:criterion}.}
\label{tab:service}
\end{table}

\begin{table*}[t]
\centering\small
\begin{tabular*}{\textwidth}{@{\extracolsep{\fill}}llrrrrrr}
\toprule
Axis / size & Corpus & Conf. & Fixed@$\bar t_H$ & Head & $\Delta_{\mathrm{conf}}$ & $\Delta_{\mathrm{fix}}$ & $\bar t_H$ \\
\midrule
\multirow{4}{*}{Time / 0.6B} & WikiText & 28.39 & 35.51 & 35.78 & +7.39 & +0.27$^{*}$ & 3.68 \\
 & Belebele-en & 27.14 & 34.16 & 34.03 & +6.89 & -0.14 & 3.95 \\
 & Belebele-bn & 46.05 & 53.48 & 54.20 & +8.15 & +0.71$^{*}$ & 3.78 \\
 & Sangraha-bn & 46.56 & 53.91 & 54.42 & +7.87 & +0.52$^{*}$ & 4.02 \\
\midrule
\multirow{4}{*}{Time / 1.7B} & WikiText & 40.20 & 42.77 & 43.03 & +2.83 & +0.26 & 3.78 \\
 & Belebele-en & 36.24 & 39.38 & 39.58 & +3.34 & +0.20 & 3.78 \\
 & Belebele-bn & 61.17 & 63.92 & 64.31 & +3.14 & +0.39 & 3.84 \\
 & Sangraha-bn & 59.69 & 63.04 & 63.21 & +3.52 & +0.17 & 4.02 \\
\midrule
\multirow{4}{*}{Depth / 0.6B} & WikiText & 22.25 & 22.49 & 22.46 & +0.21 & -0.03 & 3.90 \\
 & Belebele-en & 18.04 & 18.75 & 17.15 & -0.89 & -1.61$^{*}$ & 3.69 \\
 & Belebele-bn & 43.87 & 47.04 & 46.49 & +2.62 & -0.55$^{*}$ & 3.77 \\
 & Sangraha-bn & 44.25 & 46.95 & 46.63 & +2.39 & -0.31 & 4.02 \\
\bottomrule
\end{tabular*}
\caption{Next-token accuracy (\%) at a development budget of four exits. Conf.\ is confidence stopping and Head the learned head, whose realized cost on the reporting split is $\bar t_H$. Fixed@$\bar t_H$ is the best accuracy a mix of two fixed exits reaches at that cost (the upper concave envelope of the fixed-exit curve on the same split). $\Delta_{\mathrm{conf}}$ and $\Delta_{\mathrm{fix}}$ are the head's margins over these two references in points, and asterisks mark $\Delta_{\mathrm{fix}}$ intervals that exclude zero. Depth cost is a sampled-exit index.}
\label{tab:method}
\end{table*}

An allocation gap means little until we know what each group needs. Table~\ref{tab:service} splits the canonical gaps into demand and service. In eleven of twelve settings the demand gap has the same sign as the allocation gap and makes up roughly half of it or more, and under the stricter stable target the signs agree in all twelve. English word starts need more computation than continuations ($G_{\text{dem}}=-1.89$ exits on WikiText), and Bangla continuations need more than Bangla word starts ($+2.51$ on Belebele). The Bangla reversal is therefore present in what the model needs, not only in what the policy spends.

The remaining service gaps are smaller and change sign from setting to setting. At 1.7B the English service gaps are close to zero, so confidence stopping spends in proportion to need, while the same model over-serves Bangla continuations by about 1.5 exits. We draw no directional conclusion from any single service gap, because the 0.6B WikiText value changes sign as $\varepsilon$ varies (Appendix~\ref{app:eps}) and four of the twelve change sign under the stable target.

\begin{figure}[t]
\centering
\includegraphics[width=\columnwidth]{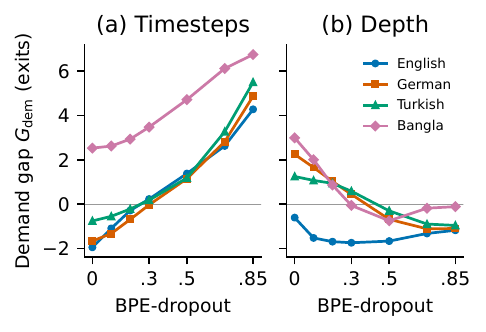}
\caption{Demand gap on the same rungs and tokens as Figure~\ref{fig:ladder}. On timesteps (a), fragmentation increases how much more computation continuation pieces need than word starts. On depth (b), through the pooled lens, it decreases it. Allocation rises on both axes, so outside Bangla the policy follows demand on timesteps and moves against it on depth.}
\label{fig:demand}
\end{figure}

On the timestep ladders, segmentation moves demand as well as spending (Figure~\ref{fig:demand}a). Within each language, fragmentation changes the demand gap by 4.6 to 6.3 exits, about as much as it changes allocation. In English, German, and Turkish, allocation follows demand closely: the correlations across rungs are $+.84$, $+.97$, and $+.99$ on the block-based ladder and $+.96$, $+.97$, and $+.98$ on the passage-aligned one. In Bangla, demand rises while allocation falls ($-.97$ and $-.71$), so the policy fails to follow a change in need. These correlations keep their signs under the stable target. Recalibrating the conversion with five further seeds moves the canonical allocation gaps by at most 0.08 exits and the demand and service gaps by at most 0.19, with no sign change (Appendix~\ref{app:conversion}).

The demand result has two limits. First, on the depth axis the same intervention moves demand the other way (Figure~\ref{fig:demand}b). Through the pooled lens the demand gap falls with fragmentation in every language: German from $+2.25$ to $-1.09$, Turkish from $+1.25$ to $-0.96$, Bangla from $+3.00$ to about zero, with English negative throughout. Fragmented continuation pieces reach their final layer-wise quality earlier, and under the stable target the policy over-serves them by up to ten exits. The layer at which a prediction forms and the timestep at which a spiking approximation settles are different quantities, and segmentation moves them in opposite directions. Second, in the small from-scratch models of Section~\ref{sec:controlled}, the tokenizer changes service rather than demand. Our finding that the gap is largely demand therefore applies to mature pretrained models on the timestep axis, and should not be read as a general property of confidence stopping.

\section{Learning to Stop}
\label{sec:head}
We train a small MLP (two hidden layers of 48 units) to decide when to stop. It sees 12 features of the trajectory so far, including confidence and entropy, their changes, running confidence statistics, and first-step values for the current and previous token, but never the target. Its label is demand: stop once $t\geq d_i$. Blocks are split three ways, for training the head, choosing its threshold, and reporting, and the threshold is chosen from 60 values to maximise development accuracy within the budget. The confidence and entropy baselines use the same protocol (Appendix~\ref{app:lens}).

The head is much better than confidence stopping (Table~\ref{tab:method}). It gains 6.89 to 8.15 points of token accuracy at 0.6B and 2.83 to 3.52 points at 1.7B, and all eight paired bootstrap intervals are above zero.

Against fixed computation the picture changes. The fair reference is the best accuracy that a mix of two fixed exits reaches at the head's own realized cost, which is the upper concave envelope of the fixed-exit accuracy curve (Figure~\ref{fig:envelope}). Against it, the head is never more than one point better, and at 1.7B no interval excludes zero. At 0.6B the English margins also depend on the calibration draw, ranging from $-0.35$ to $+0.27$ over six draws, while the Bangla margins stay between $+0.25$ and $+0.80$. On the depth axis at 0.6B the head never beats the envelope. Against confidence on depth it helps on Bangla but not on English, and the 1.7B depth probe shows the same pattern (Appendix~\ref{app:lens}). Explicit morphology features add nothing consistent.

A learned head thus recovers a lot from a weak confidence signal. Compared with fixed computation, it is at best marginally better, and not consistently so on the depth axis.

\section{Discussion and Conclusion}
Our results separate two questions: whether segmentation changes where a stopping rule spends computation, and why languages differ. The first has a clear answer. Changing segmentation, at inference time or in training, moves the allocation gap in a predictable direction on both axes and across model families. The second does not. Granularity fails as a predictor across languages, and the surprisal contrast that works better comes from the same model whose behaviour it explains. Morphology, training data, and vocabulary coverage remain entangled.

Three practical points follow for building and evaluating early-exit systems. First, report the allocation gap by word position, for each language and tokenizer, next to the mean exit, because a single average hides sign reversals. The sign of the surprisal contrast is a cheap first guess at the sign of the gap; it agreed in 23 of our 24 corpus--model pairs. Second, do not treat the gap as a bias to remove. On the timestep axis much of it is what tokens need, so measure demand before equalising anything, and name the axis and model it was measured on. Third, compare a stopping rule with the best mix of fixed exits at its realized cost, since beating confidence stopping is easy and says little.

Where a model needs computation inside a word is not fixed by the language. It depends on the segmentation, the adaptive axis, and how mature the model is, and confidence stopping follows it only some of the time.

\section*{Limitations}
\paragraph{Causal scope.} BPE-dropout changes sequence length, token identities, group membership, positions, and how familiar the input is, all at once. The passage-aligned design holds the scored characters fixed, which removes confounds from block boundaries, context spans, and missing targets, but the other factors still move together. The from-scratch models remove the distribution-shift explanation. They are, however, an order of magnitude smaller than the pretrained models and trained on far less data, cover one language pair and one tokenizer contrast, and have three seeds plus a single-seed 125.6M check. They show the direction and size of the tokenizer effect at that scale, not its mechanism. On the depth axis, any fixed lens is a fitted component between the model and the measurement. A lens fitted on canonical text under-reads fragmented input, which is why we report the pooled lens next to it.

\paragraph{Offline cost accounting.} All trajectories are computed in full, and exits are selected afterwards. Our features use only the prefix, but a model that really stopped early would have different past states and attention inputs, and could make different later predictions. We neither validate those trajectories nor measure speedups. Exit counts leave out the cost of the stopping head, lenses, unembedding, synchronization, and cache management, and depth exits are not evenly spaced in layers.

\paragraph{Generalization and statistics.} The spiking results come from one model family at two sizes. The depth probe adds two more families, but only at canonical segmentation. Calibration variation is measured (six threshold draws) and small; head-training seeds and pretrained-model seeds are not varied. Block bootstrap intervals condition on fitted policies and can understate dependence between adjacent blocks of the same document. The passage-aligned analysis resamples whole passages, which removes that dependence but still conditions on the fitted policies. The shared-slope fit is in-sample and includes languages with one observation. The German and Turkish predictions in Section~\ref{sec:failed} were recorded in our code before those ladders were run, but they were not publicly preregistered, and earlier exploration included hypotheses that failed.

\paragraph{Language measurement.} Byte-BPE space markers are only a proxy for word position. Special tokens and punctuation are not fully removed in the standard runs. Character counts are Unicode code points, not grapheme clusters or units of meaning. Token accuracy and loss depend on segmentation, so they are not directly comparable across languages. The 15-language regression is easier to interpret than the 18-language version but is not language-neutral. The gaps of English, German, and Turkish converge under heavy fragmentation, but partly in a degraded regime: the unconverted model's English loss rises from 1.12 to about 12.97 bits per character at dropout .85. Agreement between heavily fragmented English and canonical Bangla is therefore not evidence that granularity alone explains Bangla's gap.

\paragraph{The demand criterion.} Demand comes from the same model whose stopping it evaluates, so it measures internal consistency and cannot show that a group is objectively under-served. It is a first-hit index: a prediction can reach it and then get worse, so it does not bound the computation a token needs. It also depends on $\varepsilon$. Over .02 to .20, mean demand moves by about three exits while the demand gap moves much less and keeps its sign, but the WikiText service gap moves from $+0.49$ to $-0.13$, so we draw no conclusion about the direction of English service there. On the depth axis demand passes through a fitted lens, and 36--48\% of tokens never stably converge, against 5--11\% on timesteps. Demand is measured on full trajectories and describes the model as evaluated here, not a model that actually stopped early.

\paragraph{Mechanism and readouts.} We have not established a link to morphology. The boundary data available to us cover the corpora sparsely and become uninformative when a tokenizer cuts at every character (Appendix~\ref{app:morph}). The tuned-lens comparison fits one lens per corpus and splits by block rather than by document. We offer no linguistic explanation for the language intercepts or for the head's failure to transfer, since neither has been tested.

\section*{Ethical Considerations}
We analyze released model checkpoints and existing text datasets. The results concern a model and its tokenization pipeline, not the abilities of speakers or intrinsic difficulty of their languages. Lower token-level accuracy or greater simulated compute should not be used to rank language communities. We report negative results and measurement failures because overstated multilingual efficiency claims could otherwise motivate inequitable design choices. No generated text or raw corpus passages are redistributed in the manuscript package. Dataset and model terms remain applicable to reproductions. This draft was prepared with AI assistance for literature discovery, writing, and analysis tooling; responsibility for source verification, scientific claims, and the eventual submission remains with the human authors.

\paragraph{Artifacts, licenses, and compute.} We use Qwen3-0.6B and Qwen3-1.7B (Apache 2.0), mGPT-1.3B (Apache 2.0), and BLOOM-1.1B (BigScience BLOOM RAIL License v1.0), and the corpora WikiText-103 (CC BY-SA and GFDL), Belebele (CC BY-SA 4.0), and Sangraha (CC BY 4.0). All are used for research analysis consistent with their licenses. Experiments on the pretrained models ran on single NVIDIA RTX 5070 Ti (16 GB) and RTX 5090 (32 GB) GPUs, and the from-scratch models of Section~\ref{sec:controlled} were trained on a rented cloud GPU instance. We used PyTorch 2.10.0, Transformers 5.16.1, Tokenizers 0.23.2, NumPy 1.26.4, SciPy 1.14.0, and Morfessor 2.0.6.

\bibliography{references}

\clearpage
\appendix
\section{Conversion and Computational Interpretation}
\label{app:conversion}
Each spiking site wraps the input of an MLP down-projection. With channel threshold $\theta$ and burst bound $K=8$, input $x_t$ is decomposed as $\tilde x_t=\operatorname{clip}(x_t,-K\theta,K\theta)$ and $r_t=x_t-\tilde x_t$. Membrane update and output are
\begin{align}
 u_t&=v_{t-1}+\tilde x_t,\\
 n_t&=\operatorname{clip}(\operatorname{trunc}(u_t/\theta),-K,K),\\
 v_t&=u_t-\theta n_t,\qquad s_t=\theta n_t+r_t.
\end{align}
The initial membrane is $v_0=\theta/2$. The source implementation uses truncation toward zero, not a floor for negative values. The default readout passes the instantaneous spike-encoded activation and averages logits across elapsed timesteps. Threshold calibration disables spiking and averages per-batch channel percentiles. It is repeated once per model setup, then frozen across a segmentation ladder.

All unconverted Transformer operations and sparse residuals remain analog. The simulations execute dense GPU kernels. A reduction in selected timesteps is therefore only a policy-level cost proxy. The manuscript does not reuse earlier repository energy claims as a speedup result for the learned head. The evaluation holds the full token block fixed across simulation timesteps and resets neuron state between blocks, making this a specific hybrid approximation experiment rather than a native autoregressive spiking architecture.

\begin{figure}[h]
\centering
\includegraphics[width=\columnwidth]{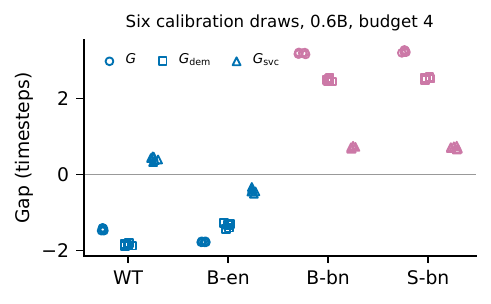}
\caption{Allocation, demand, and service gaps of the 0.6B model at budget four under six threshold-calibration draws (the main run and five re-conversions on different WikiText blocks), evaluation blocks held fixed. Each mark is one draw; the six draws are almost superimposed, and no gap changes sign.}
\label{fig:calib}
\end{figure}

Calibration is the one fitted component of the timestep axis. Figure~\ref{fig:calib} shows the six calibration draws of Section~\ref{sec:service}: allocation gaps move by at most $0.08$ and demand and service gaps by at most $0.19$ across draws.

\section{Depth Probe and Policy Details}
\label{app:lens}

\begin{figure}[t]
\centering
\includegraphics[width=\columnwidth]{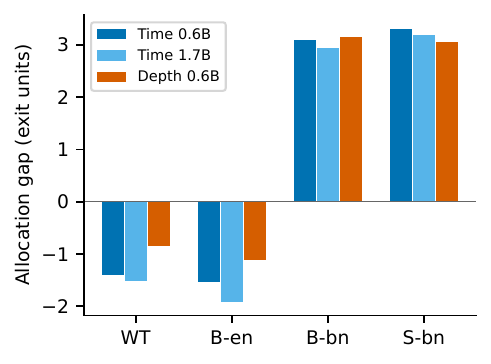}
\caption{Confidence allocation signs recur across two converted model sizes and the non-spiking tuned-lens probe. Depth values are sampled-exit indices; their magnitudes are not directly comparable hardware costs. WT: WikiText; B: Belebele; S: Sangraha.}
\label{fig:axes}
\end{figure}

\begin{figure*}[t]
\centering
\includegraphics[width=\textwidth]{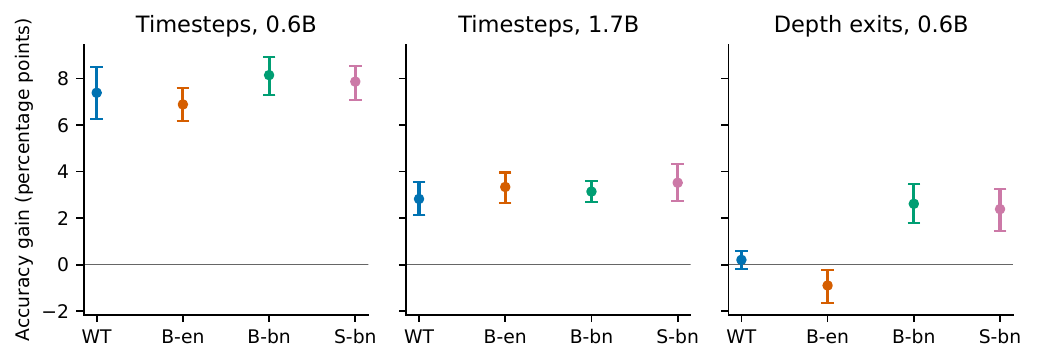}
\caption{Learned-head gains over confidence stopping at development budget four. Bars show archived 95\% paired block-bootstrap intervals, conditional on the fitted head and threshold. Gains reproduce across SNN sizes but not uniformly on depth. WT: WikiText; B-en/B-bn: Belebele English/Bangla; S-bn: Sangraha Bangla.}
\label{fig:method}
\end{figure*}

The 0.6B Transformer has 28 layers. Tuned lenses attach to hidden-state indices $1,3,5,7,8,10,12,14,16,18,20,21,23,25,27$. The final exit uses the model's actual output. Each lens has an identity-initialized affine transformation, followed by the final norm and unembedding. Optimization uses final-distribution KL targets, 800 AdamW updates, batch size 256, and learning rate $.001$. Each corpus supplies 32 additional training blocks. Hidden states are cached in half precision and readout calculations use single precision.

The evaluation stream precedes the lens training blocks in the same deterministically shuffled corpus. This prevents reuse of evaluation tokens as lens-training tokens, but a document can cross the block boundary. Corpus-specific and shared-English lens configurations are distinct experiments; the former is used in the manuscript. The last output is exact in both. A median first-hit index is not a stable convergence certificate. An English-trained shared lens transfers poorly to Bangla in the archived runs, which is why each corpus gets its own lens, although this adds another fitted component that differs across the comparison. The untrained logit lens gives very weak shallow predictions and a different allocation pattern; we treat it as a measure of readout sensitivity. The Qwen3-1.7B depth probe gives allocation gaps of $-0.04$ ($[-0.26,+0.19]$), $-0.92$, $+2.86$, and $+3.00$ on WikiText, Belebele-English, Belebele-Bangla, and Sangraha. On that probe the learned head's margins over confidence stopping are $-1.14$, $-0.94$, $+2.91$, and $+2.82$ points, and over the fixed-exit envelope $-0.52$, $-1.28$, $+0.73$, and $+0.90$, with the second and third intervals excluding zero. None of the depth experiments implements a deployed early-exit decoder with missing past-layer states.

The learned stopping head has 3,025 trainable parameters with 12 inputs: current confidence and entropy, their one-step changes, running confidence mean and maximum, deviation from the running mean, normalized timestep, the current token's first-step confidence and entropy, and those first-step values for the previous token (zeroed across block boundaries). It is trained with AdamW for 400 epochs at learning rate $.003$ and weight decay $.0001$. Its features are standardized on the training split. Entropy is divided by $\log(151936)$ in the archived implementation, a fixed vocabulary-output normalization. The head consumes trajectory summaries rather than hidden states. Training indices satisfy block index modulo three equal to zero, development indices equal to one, and reporting indices equal to two. The 48-block corpora provide 16 blocks per split (8,176 reporting tokens); Sangraha provides 10 reporting blocks (5,110 tokens). No formal noninferiority margin to final-compute quality was prespecified.

For nominal budgets 2, 4, and 6, threshold selection maximizes development accuracy subject to its average exit budget. At a fixed budget, the confidence baseline can be a weaker allocator than a constant exit. We report its comparison because it isolates the effect of changing the stopping signal, not because it exhausts the efficient-inference literature. The label-informed morphology floor should be interpreted only as a diagnostic selection rule. Corpus transfer tables in the repository use rounded saved thresholds and can have unequal destination costs; they are not used as evidence of exactly compute-matched transfer in this paper.

\begin{figure}[h]
\centering
\includegraphics[width=\columnwidth]{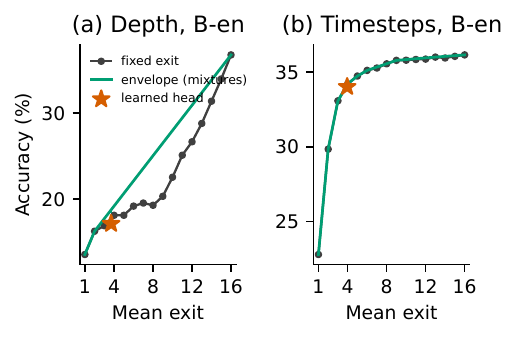}
\caption{Why the fixed-exit reference must be the envelope. Reporting-split accuracy at each fixed exit (dots), the upper concave envelope reachable by mixing two fixed exits (line), and the learned head at its realized cost (star), Belebele-English at 0.6B. (a) The tuned-lens depth curve is strongly convex over the first eight exits, so the envelope at the head's cost lies a full point above the adjacent-exit interpolation and the head falls below both. (b) The timestep curve is concave, so the two references coincide.}
\label{fig:envelope}
\end{figure}

Figure~\ref{fig:envelope} shows the two nonadaptive references of Section~\ref{sec:head} on one corpus. The interpolation and envelope values for every setting are both archived; the manuscript reports the envelope.

\section{Complete Intervention Summary}
\label{app:ladder}
\begin{table*}[t]
\centering\small
\begin{tabular}{lrrrrrrr}
\toprule
Language & $p=0$ & $.1$ & $.2$ & $.3$ & $.5$ & $.7$ & $.85$\\
\midrule
English & $-1.44$ & $-0.45$ & $+0.44$ & $+1.60$ & $+3.35$ & $+3.42$ & $+3.37$\\
German & $-4.00$ & $-3.93$ & $-3.66$ & $-2.97$ & $-0.04$ & $+2.58$ & $+2.96$\\
Turkish & $-5.03$ & $-4.73$ & $-4.37$ & $-3.56$ & $-1.42$ & $+1.38$ & $+2.56$\\
Bangla & $+3.15$ & $+3.03$ & $+2.97$ & $+2.77$ & $+2.20$ & $+1.80$ & $+1.99$\\
\bottomrule
\end{tabular}
\caption{Mean allocation gap across segmentation draws at each dropout level. Only nonzero rungs have three draws. The near-zero German $.5$ rung is the only one with draw-level sign disagreement.}
\end{table*}
The canonical English ladder scores 13,797 targets from 137 selected source passages; Bangla also scores 13,797 targets but from 27 passages. These differing source counts follow canonical-token-budget selection, so the block-based ladders are not evaluated on paired content; the passage-aligned ladder of Table~\ref{tab:aligned} is. Draw-specific token counts and decoded character totals are preserved in the small evidence files accompanying the manuscript. Canonical cells are not treated as three independent repetitions.

\begin{table*}[t]
\centering\small
\begin{tabular}{llrrrrrrrr}
\toprule
 & Language & $p=0$ & $.1$ & $.2$ & $.3$ & $.5$ & $.7$ & $.85$ & $r$\\
\midrule
\multirow{4}{*}{$G$}
 & English & $-1.49$ & $-0.66$ & $+0.30$ & $+0.89$ & $+2.45$ & $+3.46$ & $+3.44$ & $+0.96$\\
 & German & $-4.38$ & $-4.39$ & $-4.21$ & $-3.81$ & $-1.13$ & $+2.46$ & $+3.11$ & $+0.97$\\
 & Turkish & $-5.35$ & $-5.17$ & $-5.08$ & $-4.32$ & $-2.11$ & $+1.40$ & $+2.69$ & $+0.98$\\
 & Bangla & $+3.26$ & $+3.10$ & $+2.97$ & $+2.81$ & $+2.42$ & $+2.48$ & $+2.79$ & $-0.71$\\
\midrule
\multirow{4}{*}{$G_{\text{dem}}$}
 & English & $-1.94$ & $-1.09$ & $-0.28$ & $+0.23$ & $+1.38$ & $+2.63$ & $+4.28$ & $+0.96$\\
 & German & $-1.66$ & $-1.32$ & $-0.68$ & $-0.03$ & $+1.15$ & $+2.82$ & $+4.87$ & $+0.97$\\
 & Turkish & $-0.75$ & $-0.55$ & $-0.21$ & $+0.18$ & $+1.19$ & $+3.27$ & $+5.50$ & $+0.98$\\
 & Bangla & $+2.53$ & $+2.63$ & $+2.94$ & $+3.48$ & $+4.72$ & $+6.12$ & $+6.75$ & $-0.71$\\
\bottomrule
\end{tabular}
\caption{Passage-aligned intervention: allocation gap $G$ and demand gap $G_{\text{dem}}$ at each dropout level, with the Pearson correlation $r$ between them across the seven rungs. The 60 scored Belebele passages are the same \texttt{link} identifiers in all four languages, and the evaluated character total is identical at every rung of a language. Demand rises with fragmentation everywhere; allocation follows it except in Bangla.}
\label{tab:aligned}
\end{table*}
The passage-aligned run uses the same conversion, thresholds, budget, and dropout rungs as the block-based ladder, and differs only in that each passage is scored as its own sequence with a prepended end-of-sequence marker. That marker makes the first character of every passage a scored target, which is what equalizes character coverage across rungs; without it, coverage differs by about half a percent because the first token of a passage is never predicted. Decoded Bangla characters exceed the source count by 1.1\%. The cause is Unicode normalization rather than the segmentation: the tokenizer normalizes to NFC, and the Bengali letters YYA and RRA are composition exclusions, so each decomposes into a base letter plus a nukta sign and adds one character. On the 60 shared passages this adds 320 characters and removes 4 where a vowel-sign pair composes, a net of 316. The offset is a constant per language, so it shifts Bangla's characters-per-token and bits-per-character denominators by about 1.1\% and cancels in every across-rung comparison.

The bare-marker exclusion is applied after selecting stopping points and removes only vocabulary pieces equal to the isolated space or newline marker. It does not remove all punctuation or establish which token is the first lexical character after an isolated space. More complete character-offset labeling would improve interpretability. Reported within-draw intervals resample blocks while holding the threshold fixed; they do not include threshold-estimation uncertainty.

\begin{table*}[t]
\centering\small
\begin{tabular}{llrrrrrrrr}
\toprule
 & Language & $p=0$ & $.1$ & $.2$ & $.3$ & $.5$ & $.7$ & $.85$ & $r$\\
\midrule
\multirow{4}{*}{\shortstack[l]{$G$\\ canonical lens}}
 & English & $-0.88$ & $-0.47$ & $-0.18$ & $+0.05$ & $+0.45$ & $+0.60$ & $+0.51$ & $-0.86$\\
 & German & $-3.97$ & $-4.01$ & $-3.99$ & $-3.78$ & $-3.13$ & $-1.88$ & $-0.53$ & $-0.96$\\
 & Turkish & $-5.45$ & $-5.48$ & $-5.48$ & $-5.42$ & $-5.01$ & $-3.82$ & $-2.20$ & $-0.97$\\
 & Bangla & $+3.13$ & $+2.95$ & $+2.72$ & $+2.54$ & $+2.27$ & $+2.17$ & $+2.22$ & $+0.98$\\
\midrule
\multirow{4}{*}{\shortstack[l]{$G$\\ pooled lens}}
 & English & $-0.54$ & $+0.92$ & $+1.93$ & $+2.73$ & $+3.51$ & $+3.60$ & $+3.48$ & $-0.55$\\
 & German & $-3.45$ & $-3.51$ & $-3.38$ & $-2.90$ & $-0.14$ & $+3.02$ & $+3.32$ & $-0.90$\\
 & Turkish & $-5.17$ & $-5.25$ & $-5.26$ & $-4.99$ & $-3.51$ & $+1.18$ & $+2.98$ & $-0.92$\\
 & Bangla & $+3.04$ & $+2.89$ & $+2.75$ & $+2.50$ & $+1.87$ & $+2.68$ & $+3.06$ & $+0.63$\\
\midrule
\multirow{4}{*}{\shortstack[l]{$G_{\text{dem}}$\\ pooled lens}}
 & English & $-0.60$ & $-1.52$ & $-1.69$ & $-1.74$ & $-1.66$ & $-1.32$ & $-1.18$ & $-0.55$\\
 & German & $+2.25$ & $+1.66$ & $+1.03$ & $+0.44$ & $-0.68$ & $-1.11$ & $-1.09$ & $-0.90$\\
 & Turkish & $+1.25$ & $+1.08$ & $+0.94$ & $+0.59$ & $-0.30$ & $-0.89$ & $-0.96$ & $-0.92$\\
 & Bangla & $+3.00$ & $+2.01$ & $+0.87$ & $-0.06$ & $-0.75$ & $-0.18$ & $-0.11$ & $+0.63$\\
\bottomrule
\end{tabular}
\caption{Depth-axis ladder on the same tokens as Table~\ref{tab:aligned}, through two readouts, each fitted once and frozen across all rungs: a lens fitted on canonical text only, and one fitted on the pooled support of the ladder. $G$ is the allocation gap at budget four of 16 exits and $G_{\text{dem}}$ the first-hit demand gap; $r$ is their correlation across the seven rungs. Under the pooled lens the allocation curves match the timestep ladder; under both, the demand gap moves the other way.}
\label{tab:depthladder}
\end{table*}

\paragraph{Depth-axis ladder.} The depth ladder of Section~\ref{sec:depthladder} reads the realized token sequences of the passage-aligned run from its archive, so its 76 cells have exactly the target tokens of Table~\ref{tab:aligned} (verified cell by cell). Each language's lens is fitted on the shared Belebele passages that follow the 60 evaluated ones, canonically segmented, until 24{,}000 tokens are collected (246 passages for English), with the hyperparameters of Appendix~\ref{app:lens}, then frozen. The longest scored sequence is 2{,}354 tokens (Bangla at dropout $.85$, which averages about 1{,}200 tokens per passage; English, German, and Turkish stay under 1{,}050), below the 2{,}400-token cap, so no passage is truncated on either axis. Under fragmentation the frozen lens loses shallow-exit quality: the first-exit NLL of English word-start pieces is $6.6$ nats at canonical segmentation and $9.1$ at dropout $.85$, whereas the first spiking timestep's is $5.6$ and $2.3$, and the share of tokens that never stably converge falls from 48\% to 20\% because heavily fragmented pieces are trivially predictable at every layer. The pooled-support lens is fitted on the same held-out passages segmented at dropout $0$, $.3$, and $.85$ with equal token shares (24{,}482 tokens for English), with the same hyperparameters, and is likewise frozen across all 76 cells. It restores shallow-exit quality on fragmented input (first-exit NLL of English word-start pieces $2.2$ nats at dropout $.85$) at a small cost on canonical input ($6.8$ against $6.6$), and the never-stably-converged share then falls from 48\% to 19\% across the English ladder. Under it the allocation curves of Table~\ref{tab:depthladder} lie within two exits of the timestep ladder at every rung and within one exit at 23 of 28; the largest departures are English's early rungs, where the depth response is faster (it crosses zero between dropout $0$ and $.1$ rather than between $.1$ and $.2$). The demand curves keep the sign pattern of the canonical lens with smaller first-hit ranges (1.1--3.7 exits) and larger stable-target ranges (5.4--7.4). A lens refitted at every rung, which we do not report, would let the readout absorb part of the treatment and is not a valid instrument for this question.

\paragraph{Lens-fitting seed and budget.} Refitting the pooled lens under two further seeds (2345 and 3456, changing the initialization and the training batches, with the same held-out passages) and rescoring the canonical and dropout-$.85$ cells gives endpoints of $-0.55\to+3.50$ and $-0.52\to+3.49$ for English, $-3.28\to+3.39$ and $-3.61\to+3.26$ for German, $-5.14\to+3.13$ and $-5.20\to+3.02$ for Turkish, and $+3.02\to+3.12$ and $+3.01\to+3.04$ for Bangla, against $-0.54\to+3.48$, $-3.45\to+3.32$, $-5.17\to+2.98$, and $+3.04\to+3.06$ under the reported lens: no endpoint moves by more than $0.17$ exits, and the first-hit demand gap falls between the two rungs in all four languages under all three lenses (Bangla's dropout-$.85$ value is $-0.11$, $+0.62$, and $+0.11$, so ``about zero''). The 32 rescored cells use the same realized tokens as the full ladder. Re-reading both depth ladders at budgets 2, 6, and 8 of 16 exits, English, German, and Turkish cross zero under the pooled lens at every budget from 4 upward (at budget 2 the ranges compress to 0.7--2.8 exits and German and Turkish still cross), Bangla stays positive through budget 6, and the allocation--demand correlation keeps its sign in every language at every budget ($-0.28$ to $-0.92$ for English, German, and Turkish; $+0.63$ to $+0.89$ for Bangla).

\section{Interpolation and Regression Scope}
\label{app:convergence}
\begin{figure}[t]
\centering
\includegraphics[width=\columnwidth]{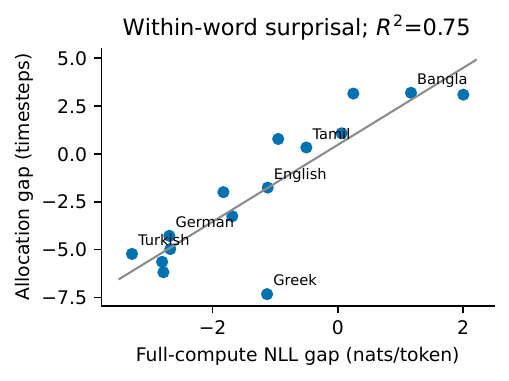}
\caption{Full-compute surprisal contrast predicts allocation across the 15 whitespace-delimited languages. Both axes are derived from the same model, so the relationship describes model behavior rather than establishing an independent linguistic cause.}
\label{fig:info}
\end{figure}

\begin{figure}[h]
\centering
\includegraphics[width=\columnwidth]{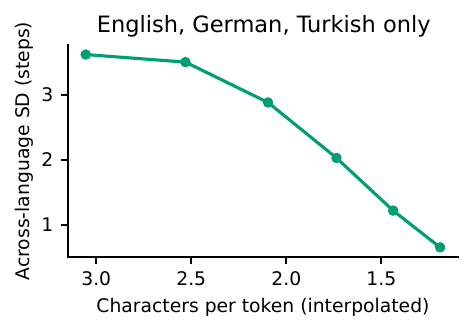}
\caption{Descriptive interpolation over the overlapping ranges of English, German, and Turkish. The spread decreases by 5.54-fold. Bangla is excluded by range support, and interpolation adds no new observations.}
\end{figure}
The shared-slope model of Section~\ref{sec:info} is
\begin{equation}
 G_{\ell,r}=\alpha_\ell+\beta[-\log g_{\ell,r}]+\epsilon_{\ell,r},
\end{equation}
where $g$ is characters per token, fitted to 15 canonical cross-language points and 28 draw-averaged ladder rungs; the shared slope is $\beta=4.61$. Flexible intercepts absorb variation by design, and the improvement is not held-out predictive performance. The common-granularity curve interpolates allocation linearly in negative log characters per token. Each displayed point requires at least three supported ladders; in the archived result all six points have exactly three. The endpoints have standard deviations 3.6196 and 0.6538. No confidence interval is assigned to the ratio, and the underlying high-dropout quality degradation remains relevant.

The 15-language information analysis includes English, German, Swahili, Turkish, Hungarian, Finnish, Russian, Arabic, Hebrew, Greek, Hindi, Amharic, Bangla, Tamil, and Telugu. Chinese, Japanese, and Burmese are retained only in the explicitly labeled 18-language sensitivity calculation. Negative log characters per token and log tokens per morpheme differ by a language-specific constant if a morphology estimate is held fixed. Language intercepts absorb that constant, so fitting both axes is not independent morphological evidence.

\section{Tolerance Sensitivity of the Demand Decomposition}
\label{app:eps}
\begin{table}[h]
\centering\small
\begin{tabular}{lrrrr}
\toprule
Corpus & $\varepsilon$ & mean $d$ & $G_{\text{dem}}$ & $G_{\text{svc}}$\\
\midrule
\multirow{4}{*}{WikiText}
 & $.02$ & 7.17 & $-1.91$ & $+0.49$\\
 & $.05$ & 6.25 & $-1.89$ & $+0.47$\\
 & $.10$ & 5.31 & $-1.67$ & $+0.25$\\
 & $.20$ & 4.27 & $-1.29$ & $-0.13$\\
\midrule
\multirow{4}{*}{Belebele-en}
 & $.02$ & 6.97 & $-1.32$ & $-0.44$\\
 & $.05$ & 6.07 & $-1.43$ & $-0.33$\\
 & $.10$ & 5.14 & $-1.34$ & $-0.42$\\
 & $.20$ & 4.11 & $-1.06$ & $-0.70$\\
\midrule
\multirow{4}{*}{Belebele-bn}
 & $.02$ & 5.24 & $+2.81$ & $+0.38$\\
 & $.05$ & 4.39 & $+2.51$ & $+0.68$\\
 & $.10$ & 3.61 & $+2.08$ & $+1.11$\\
 & $.20$ & 2.83 & $+1.53$ & $+1.66$\\
\midrule
\multirow{4}{*}{Sangraha-bn}
 & $.02$ & 5.15 & $+2.85$ & $+0.42$\\
 & $.05$ & 4.30 & $+2.52$ & $+0.75$\\
 & $.10$ & 3.54 & $+2.07$ & $+1.20$\\
 & $.20$ & 2.76 & $+1.50$ & $+1.78$\\
\bottomrule
\end{tabular}
\caption{Demand and service gaps for the 0.6B timestep model as the tolerance $\varepsilon$ of Equation~\ref{eq:demand} varies. The allocation gap $G$ does not depend on $\varepsilon$. Mean demand moves by roughly three exits; the demand gap moves far less and never changes sign. The service gap changes sign only on WikiText.}
\end{table}
The tolerance controls how close to final quality a prediction must come before its demand is considered met, so a larger $\varepsilon$ mechanically lowers every demand and raises every service value. What matters for Section~\ref{sec:service} is whether the between-group contrast is stable, and the demand contrast is: it keeps its sign and most of its magnitude throughout. Because $G=G_{\text{dem}}+G_{\text{svc}}$ and $G$ is fixed, any movement in the demand gap appears in the service gap, which is why the smaller residual is the more fragile of the two components.

\section{First-Hit versus Stable Demand}
\label{app:criterion}

Equation~\ref{eq:demand} defines demand as the \emph{first} exit within $\varepsilon$ of the final loss. A prediction can leave that band again afterwards, so the target does not certify that the model is finished. We therefore recompute the decomposition with a stricter target, the first exit from which the loss \emph{stays} within the band,
\[
d^{\text{stab}}_i=\min\bigl(\{t:\ell_{i,t'}\le\ell_{i,T}+\varepsilon\ \ \forall t'\ge t\}\bigr),
\]
which is one past the last excursion and is never smaller than the first-hit index. The policy and budget are unchanged; only the reference moves.

\begin{table}[h]
\centering\small\setlength{\tabcolsep}{3.5pt}
\begin{tabular}{llrrrr}
\toprule
& & \multicolumn{2}{c}{first hit} & \multicolumn{2}{c}{stable}\\
\cmidrule(lr){3-4}\cmidrule(lr){5-6}
Setting & Corpus & $G_{\text{dem}}$ & $G_{\text{svc}}$ & $G_{\text{dem}}$ & $G_{\text{svc}}$\\
\midrule
\multirow{4}{*}{Time / 0.6B}
 & WT    & $-1.89$ & $+0.47^{*}$ & $-2.33$ & $+0.91^{*}$\\
 & B-en  & $-1.43$ & $-0.33^{*}$ & $-1.71$ & $-0.05$\\
 & B-bn  & $+2.51$ & $+0.68^{*}$ & $+3.87$ & $-0.68^{*}$\\
 & S-bn  & $+2.52$ & $+0.75^{*}$ & $+3.94$ & $-0.67^{*}$\\
\midrule
\multirow{4}{*}{Time / 1.7B}
 & WT    & $-1.56$ & $-0.05$ & $-2.59$ & $+0.98^{*}$\\
 & B-en  & $-1.86$ & $-0.07$ & $-2.44$ & $+0.52^{*}$\\
 & B-bn  & $+1.40$ & $+1.52^{*}$ & $+2.41$ & $+0.52^{*}$\\
 & S-bn  & $+1.53$ & $+1.59^{*}$ & $+2.47$ & $+0.65^{*}$\\
\midrule
\multirow{4}{*}{Depth / 0.6B}
 & WT    & $+0.68$ & $-1.38^{*}$ & $-0.10$ & $-0.60^{*}$\\
 & B-en  & $-0.60$ & $-0.66^{*}$ & $-0.58$ & $-0.68^{*}$\\
 & B-bn  & $+3.28$ & $-0.14$ & $+4.17$ & $-1.03^{*}$\\
 & S-bn  & $+3.38$ & $-0.32^{*}$ & $+4.18$ & $-1.12^{*}$\\
\bottomrule
\end{tabular}
\caption{The decomposition under both demand targets, $\varepsilon=0.05$, budget four. Asterisks mark cluster-bootstrap intervals excluding zero. The demand gap matches the allocation sign in all twelve settings under the stable target, including the depth WikiText case that is the exception under first hit. Four service gaps change sign between targets.}
\label{tab:criterion}
\end{table}

Two things follow from Table~\ref{tab:criterion}, and they point in opposite directions. The claim that demand carries the allocation gap is \emph{strengthened}: under the stable target the demand gap matches the allocation sign in twelve of twelve settings rather than eleven, and the depth WikiText exception named in Table~\ref{tab:service} disappears. The claim that any particular service gap has a particular direction is \emph{weakened}: four of twelve change sign, and on the 0.6B Bangla corpora the change is from $+0.68$ and $+0.75$ to $-0.68$ and $-0.67$ with both intervals excluding zero. A confidently signed but target-dependent residual is not evidence about which group is under-served.

Mean demand rises substantially under the stricter target, from 6.25 to 8.60 exits on WikiText at 0.6B and from 6.13 to 12.29 on Sangraha in the depth probe. On the depth axis 36--48\% of tokens never stably converge and are pinned at $T$, against 5--11\% on the timestep axis. Per-layer losses under a tuned lens are not monotone, so stable convergence is a demanding criterion there; this censoring is a further reason to treat the depth demand values as the more fragile of the two axes.

\paragraph{The ladders under the stable target.} Recomputing the decomposition of every ladder rung with $d^{\text{stab}}$ leaves the timestep-axis conclusions of Section~\ref{sec:service} in place. On the block-based ladder the stable demand gap runs from $-1.17$ to $+8.28$ in English, $-1.80$ to $+7.96$ in German, $-0.77$ to $+7.92$ in Turkish, and $+3.92$ to $+6.27$ in Bangla, with allocation--demand correlations $+.88$, $+.97$, $+.99$, and $-.97$; on the passage-aligned ladder the corresponding correlations are $+.96$, $+.97$, $+.98$, and $-.66$. The only qualitative change is that Bangla's stable-target demand stops rising at dropout $.7$ ($6.33$ then $6.27$ block-based, $6.33$ then $6.34$ aligned), so ``demand rises with fragmentation'' holds strictly for Bangla under the first-hit target and up to a plateau under the stable one. Service gaps change sign between targets at several rungs, as on the standard caches, and no directional service claim is made for any single rung.

\section{Tokenizer-Controlled Models}
\label{app:controlled}
\begin{table}[h]
\centering\footnotesize\setlength{\tabcolsep}{2.5pt}
\begin{tabular}{lrrrr}
\toprule
& \multicolumn{3}{c}{B$-$A by seed} & \\
\cmidrule(lr){2-4}
& 0 & 1 & 2 & mean (sd)\\
\midrule
Bangla $\Delta G$  & $-6.00$ & $-6.27$ & $-6.06$ & $-6.11$ ($0.14$)\\
English $\Delta G$ & $+0.27$ & $+0.15$ & $-0.48$ & $-0.02$ ($0.40$)\\
Bangla $\Delta G$, eq.\ tok. & $-6.11$ & -- & -- & --\\
Bangla $\Delta G_{\text{dem}}^{\text{s}}$ & $-0.39$ & $-0.49$ & $-0.57$ & $-0.48$ ($0.09$)\\
Bangla $\Delta G_{\text{svc}}^{\text{s}}$ & $-5.62$ & $-5.78$ & $-5.49$ & $-5.63$ ($0.14$)\\
\midrule
Bangla $G$, Arm A & $+0.93$ & $+0.79$ & $+0.87$ & $+0.86$ ($0.07$)\\
Bangla $G$, Arm B & $-5.08$ & $-5.48$ & $-5.20$ & $-5.25$ ($0.20$)\\
English $G$, Arm A & $-0.94$ & $-0.83$ & $-0.29$ & $-0.69$ ($0.35$)\\
English $G$, Arm B & $-0.66$ & $-0.68$ & $-0.76$ & $-0.70$ ($0.05$)\\
\bottomrule
\end{tabular}
\caption{From-scratch 49.1M models under two tokenizers differing only in Bangla exposure, budget 6.5 of 13 exits, 300 parallel Belebele passages. Seeds pair document order and initialization across arms; superscript s marks the stable demand target. Three seeds are descriptive; no interval over seeds is reported. Within-seed passage-cluster intervals on $G$ have half-widths of $0.06$--$0.15$.}
\label{tab:controlled}
\end{table}

\paragraph{Tokenizers.} Both arms are byte-level BPE with the same 16{,}000-piece vocabulary, trained on 20M English characters from the WikiText-103 training split plus 4{,}000 (Arm A) or 10M (Arm B) Bangla characters from Sangraha. The obvious manipulation, an English-heavy 90/10 tokenizer corpus against a balanced one, does not work: fertility is a cliff in the second language's character budget, and a few thousand characters already buy most of the subword inventory (Bangla characters per token $0.40$, $0.78$, $0.93$, $1.09$, $1.15$, $1.31$, and $1.40$ at $0$, $1.4$k, $4$k, $10$k, $20$k, $350$k, and $10$M Bangla characters), so 90/10 moves Bangla by $1.03\times$. The induction algorithm is no better as a lever: WordPiece is indistinguishable from BPE ($1.001\times$) and Unigram moves English more than Bangla. The arms are therefore $4$k against $10$M characters.

\paragraph{Training and evaluation.} Each arm trains a 12-layer, width-512 pre-norm decoder with rotary positions and tied embeddings (49.1M parameters) for one epoch over the same documents, selected by character budget before tokenization so that selection cannot depend on the tokenizer, joined by end-of-sequence markers and cut into 1{,}024-token windows so that no text is dropped. Because the same text is more tokens under Arm A, the arms consume 386M and 278M tokens and correspondingly different compute; equal documents, equal tokens, and equal compute are three different comparisons, and the matched-token row of Table~\ref{tab:controlled} compares checkpoints at 262M tokens for both. Readouts are per-layer affine lenses fitted, per language, on documents held out from training and disjoint from evaluation, then frozen; a readout fitted on English alone was found to harm Bangla enough to invert a conclusion. The stopping threshold is chosen per model to hit the target mean exit, so arms are compared at matched compute per token. The 125.6M pair (16 layers, width 768, seed 0, 16 exits) gives Bangla $\Delta G=-7.61$ and $-7.09$ at budgets 8.5 and 4.25 of 17, demand shares 5.9\% and 6.3\%, and an English control of $-0.75$ at both budgets, outside the 49.1M seed range; its evaluation record survives only as a console log because the compute instance was terminated before the result files were retrieved, and it is reported for that reason as a single-seed check rather than as a result.

\begin{figure}[h]
\centering
\includegraphics[width=\columnwidth]{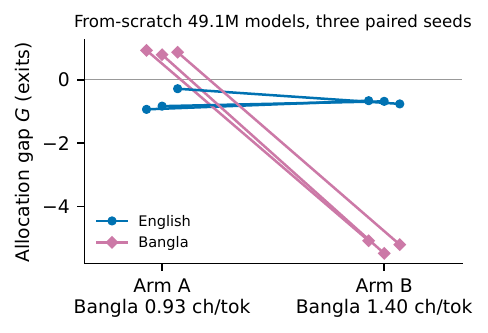}
\caption{Allocation gap of each from-scratch model, with the two arms of a seed joined by a line. The three Bangla lines fall by six exits in parallel; the three English lines are flat and cross.}
\label{fig:controlled}
\end{figure}

\paragraph{Caveats.} Figure~\ref{fig:controlled} shows the per-seed pairing. Bangla continuation share moves from 86.2\% to 79.2\% between arms, so group composition changes with the treatment. The English never-stably-converged share differs sharply between arms (9.9\% against 27.8\%), making the English stable-demand figure the weakest number here. The exits are depth exits, not spiking timesteps, and there is one language pair.

\section{Morphological Nulls and Preprocessing}
\label{app:morph}
The repository explored Morfessor segmentation \citep{morfessor} and UniMorph-derived morphology resources \citep{unimorph}. Morfessor-derived fragmentation is strongly collinear with character granularity, so it does not identify morphological alignment independently. The second analysis compares tokenizer cuts with boundaries inferred from annotated lemma--form pairs, and explicit segmentation resources where available. The lemma--form proxy is not equivalent to gold morpheme segmentation for every language.

To control the mechanical tendency of dense tokenization to hit more boundaries, the analysis uses a chance correction. If a word has $L$ characters and $C$ distinct interior cut positions, a uniformly placed cut set hits a particular boundary with probability $C/(L-1)$. Observed hits divided by their expected count define lift. The archived cross-language lift relationships are near zero (reported $R^2<.02$ for onset NLL, surprisal gap, and allocation gap). Sparse in-corpus coverage and fully saturated cut sets make these tests weak or degenerate in some languages. In particular, lift fixed at one under cuts at every interior position is uninformative, not evidence against morphology itself.

An earlier extraction routine used Python's regular-expression word class, which omits combining marks. It fragmented words at vowel signs in several scripts. The current morphology code splits on whitespace and trims edge punctuation using Unicode categories. We exclude any earlier, uncorrected morphology output from the positive argument; the primary allocation calculations classify token spellings directly and do not use that word extractor. Resolving this bug does not establish a morphological explanation of the allocation intercepts.

\end{document}